%% file: main.tex
\documentclass[9pt,twocolumn,twoside]{srformat}
\usepackage{stfloats}
\usepackage{float}
\usepackage{dsfont}
\usepackage{bbm}
\usepackage{csquotes}
\usepackage[maxnames=99, sorting=none]{biblatex}
\usepackage[normalem]{ulem}
\usepackage{pifont}
\usepackage{makecell}
\usepackage{amssymb}
\usepackage{amsthm}
\usepackage{multirow}
\usepackage{array}
\usepackage{adjustbox} % for adjustbox to adjust images
\usepackage{wrapfig}

\usepackage{framed}
\usepackage{tikz}
\usepackage{amsmath}
\usepackage{pifont}
\usepackage{bm}
\usepackage[T1]{fontenc}    % use 8-bit T1 fonts
\usepackage{hyperref}       % hyperlinks
\usepackage{url}            % simple URL typesetting
\usepackage{booktabs}       % professional-quality tables
\usepackage{amsfonts}       % blackboard math symbols
\usepackage{nicefrac}       % compact symbols for 1/2, etc.
\usepackage{xcolor}         % colors

 \usepackage{graphicx}
\usepackage{colortbl}
\usepackage{color}
\usepackage{enumitem}
\setlist{nolistsep}
\usepackage{xcolor}
\usepackage{diagbox}
\usepackage{makecell}
\usepackage{multirow}
\usepackage{booktabs}
\usepackage{geometry}
\usepackage{makecell}
\usepackage{hyperref}
\usepackage{booktabs}
\usepackage{array}
\usepackage{marvosym} % For \Letter symbol
\usepackage{float}
\usepackage{pdfpages}
\usepackage{appendix}
\usepackage{balance}
\input{figure_legends}

\runningtitle{ } % For use in the footer
\runningauthor{Zhicheng He \textit{et al.}}

\usepackage{newunicodechar}
\newunicodechar{ń}{\'n}

\title{Scalable next-scale autoregression for medical image generation across anatomical regions}

\author[1,$\dagger$]{Zhicheng He}
\author[1,5$\dagger$]{Yunpeng Zhao}
\author[6,$\dagger$]{Junde Wu}
\author[1,7]{Ziwei Niu}
\author[2]{Ziyue Wang}
\author[1,8,9]{Bohan Li}
\author[2]{Zijun Li}
\author[7]{Lanfen Lin}
\author[3,4]{Nan Liu}
\author[1,2,$\ast$]{Yueming Jin}

\affil[1]{Department of Biomedical Engineering, National University of Singapore, Singapore, Singapore}
\affil[2]{Department of Electrical and Computer Engineering, National University of Singapore, Singapore, Singapore}
\affil[3]{Centre for Biomedical Data Science, Duke-NUS Medical School, Singapore, Singapore}
\affil[4]{NUS Artificial Intelligence Institute, National University of Singapore, Singapore, Singapore}
\affil[5]{Department of Electrical and Computer Engineering, The University of Hong Kong, Hong Kong SAR, China}
\affil[6]{Department of Engineering, University of Oxford, Oxford, England}
\affil[7]{College of Computer Science and Technology, Zhejiang University, Hangzhou, China}
\affil[8]{School of Computer Science, Shanghai Jiao Tong University, Shanghai, China}
\affil[9]{Ningbo Institute of Digital Twin, Eastern Institute of Technology, Ningbo, Ningbo, China}
\affil[$\dagger$]{These authors contributed equally.}

\correspondingauthoraffiliation[$\ast$]{Corresponding author: Yueming Jin (ymjin@nus.edu.sg)}

\begin{abstract}
Autoregressive pretraining has been key to the scalability of large language models, yet medical generative foundation models remain predominantly based on diffusion. Here we introduce MedVAR, the first foundation model for all-round medical image generation through autoregressive training, and find it offers improved generation quality and efficiency, stronger scalability, and broader adaptability to downstream clinical tasks than diffusion-based models. A tokenizer trained on medical images and separate semantic and structural controls enable generation across six anatomical regions in computed tomography and magnetic resonance imaging. Trained on 438,905 slices from 40 datasets, including seven internal clinical centres, MedVAR generates images 12–19 times faster than 100-step diffusion baselines. Generation quality improves with model size. Pretraining on generated images improves seven-centre hepatocellular carcinoma segmentation Dice from 0.615 to 0.632, while reconstruction using MedVAR images approaches the volumetric segmentation performance of fully sampled volumes. Membership inference reaches 3.54\% sensitivity at a 1\% false-positive rate, while copy detection performs near chance. These findings establish next-scale autoregression as a scalable and versatile approach to medical image generation and downstream analysis.
\end{abstract}

\keywords{Medical image generation, next-scale autoregression, controllable generation, efficient sampling.}

\begin{document}

\maketitle
%\thispagestyle{firststyle}
%\slugnote
%\firstpagefootnote
\vspace{-13pt}% Only used for adjusting extra space in the left column of the first page

% Prefer longer paragraph endings while retaining the template geometry.
\setlength{\parfillskip}{0pt plus 0.45\linewidth}
\section{INTRODUCTION}
\input{1_introduction}

\section{RESULTS}
\input{2_results_recognition}  
\section{DISCUSSION}
\input{8_discussion}

\section{METHODS}
\input{5_method}

\enlargethispage{2\baselineskip}
\section{DECLARATIONS}
\noindent
\textbf{Conflict of interest}
The authors declare no competing interests.

\textbf{Data availability}
The use of retrospective, anonymized clinical data in this study was approved by the Ethics Committee of Sir Run Run Shaw Hospital, Zhejiang University School of Medicine (approval no. 2022-476-01), in accordance with the Declaration of Helsinki. The requirement for informed consent was waived because of the retrospective study design and the use of anonymized data. The public datasets used in this study are available from their original repositories under their respective access and licensing terms. Dataset composition and public access information are provided in Supplementary Table~1. The internal MRI data are not publicly available because of institutional governance and participant privacy restrictions; requests for access will be considered by the corresponding author and participating institutions subject to the required approvals. Source data underlying the main figures are provided with this paper.

\textbf{Code availability}
Code required to train and evaluate MedVAR and to reproduce the reported analyses will be made publicly available upon publication.

\clearpage
\twocolumn
\setlength{\parfillskip}{0pt plus 1fil}
\printbibliography[title={REFERENCES}]

\end{document}

%% file: figure_legends.tex
\newcommand{\figureonecaption}{%
\textbf{A next-scale autoregressive generator across anatomical regions.}
\textbf{a}, Anatomy and modality coverage. \textbf{b}, Distribution of 482,815 images from 40 datasets. \textbf{c}, Conditioning, generation and evaluation workflow. \textbf{d}, Seven-centre HCC MRI cohort; labels give volume counts and red contours mark tumours. \textbf{e}, FID comparisons within anatomical groups and a descriptive measure of quality and efficiency normalized to depth-30 MedVAR (Methods).%
}

\newcommand{\figuretwocaption}{%
\textbf{Generation quality, efficiency and diversity across MedVAR model sizes and generator families.}
\textbf{a}, Matched examples. \textbf{b}, Model depths. \textbf{c}, FID ranks within anatomy--modality groups using MedVAR with semantic conditions alone. \textbf{d}, FID versus latency; MedVAR depths 16 and 20 approximately match diffusion models at 330 and 600 million parameters, respectively. Diffusion curves span 10--100 steps; paired MedVAR points use semantic conditions alone or both semantic and structural conditions. \textbf{e}, Fidelity across generator families. \textbf{f}, Condition recognizability and CAS; the dashed line marks random balanced accuracy ($1/9$). \textbf{g}, Diversity under fixed conditions. N.A., not available; error bars, 95\% bootstrap CIs.%
}

\newcommand{\figurethreecaption}{%
\textbf{Medical tokenization and conditioning support explicit control.}
\textbf{a}, Natural and medical tokenizer comparison. \textbf{b}, Progressive conditioning. \textbf{c}, Conditioning ablation. \textbf{d}, Ablations of structural routing, the scale gate and the auxiliary structure head; MedVAR routes token maps of sizes $5^2$, $6^2$, $8^2$ and $10^2$. \textbf{e}, Effect of the weight applied to structural conditions. \textbf{f}, CFG and truncation ablations. Filled and hatched bars denote generation with semantic conditions alone and with both semantic and structural conditions, respectively; arrows indicate lower is better.%
}

\newcommand{\figurefourcaption}{%
\textbf{Generated images support segmentation and reconstruction of missing slices.}
\textbf{a}, HCC segmentation and reconstruction of missing slices. \textbf{b}, Dice gain for each centre over pretraining on real images; points denote matched seeds. \textbf{c}, Real images, images generated with centre labels and unconditional MedVAR images. \textbf{d}, PCA and t-SNE of RadImageNet features. \textbf{e}, Accuracy of centre prediction and maximum mean discrepancy (MMD); the dashed line is random. \textbf{f}, Matched CT reconstructions (top), pairwise signed differences for Linear--U-Net, U-Net--MedVAR and MedVAR--Real on a common scale (middle), and mean CT and MRI PSNR across volumes (bottom). Linear interpolation is deterministic, the U-Net is an independent residual baseline, and circles denote the three training seeds for the system using MedVAR. \textbf{g}, Paired 3D Dice distributions for linear interpolation, the residual U-Net, reconstruction using MedVAR and fully sampled input; boxes show medians and interquartile ranges, whiskers extend to 1.5 times the interquartile range, and points denote volumes. Arrows indicate better performance.%
}

\newcommand{\figurefivecaption}{%
\textbf{Privacy analyses and effects of data source labels.}
\textbf{a}, White-box membership inference by ROC AUC and TPR at 1\% FPR; error bars are 95\% bootstrap CIs. \textbf{b}, ROC curves at low FPR; dashed lines mark 1\% FPR and random performance. \textbf{c}, Nearest-neighbour copy detection for three sizes of the generated set. Diamonds and bars show mean $\pm$ s.d. across encoders; circles show individual encoders. \textbf{d}, Generated images and nearest training and test images with NCC, SSIM and difference maps. \textbf{e}, Stress test with and without labels identifying the data source under simulated contamination; positive interaction values indicate attenuation when the label is provided.%
}

\newcommand{\figuresixcaption}{%
\textbf{MedVAR architecture and semantic and structural conditioning.}
\textbf{a}, Coarse-to-fine token prediction with semantic and optional structural conditions; the auxiliary structure head is used only during training. \textbf{b}, Hierarchical semantic conditioning combines region and organ labels, then fuses this representation with imaging modality. A separately projected data source embedding is added at reduced weight before adaptive layer normalization. \textbf{c}, Structural attention routing embeds the organ mask and its boundary at each token-map scale. These features update the query and key projections of self-attention; the value projection is unchanged.%
}

%% file: 1_introduction.tex
Autoregressive pretraining has enabled large language models to learn from broad data collections and support diverse language tasks \cite{brown2020language}, with performance improving as model size and training resources increase \cite{kaplan2020scaling}. In medical imaging, a comparable aim is to learn from diverse scans and use that knowledge across organs, imaging modalities and analysis tasks. This is important because anatomy and image appearance vary across patients and acquisition protocols, and expert annotations are costly to obtain for each new task. Generative pretraining can learn image features that support subsequent analysis \cite{chen2020generative}. The resulting model can also supply synthetic images to expand training collections \cite{sun2025dataefficient,zhang2025genseg}. Developing this capability calls for image quality, efficient sampling and control over anatomical content within a common model.

Recent medical generative foundation models have largely developed around diffusion. Earlier adversarial methods enabled rapid synthesis \cite{nie2017medical,sun2022hierarchical}, while diffusion models expanded the range of anatomy and imaging modalities that could be generated. MedM2G \cite{zhan2024medm2g} and MINIM \cite{wang2025minim} support generation across modalities; MedDiff-FM \cite{yu2024meddiff}, MAISI \cite{guo2025maisi} and 3D MedDiffusion \cite{wang2025meddiffusion3d} extend generation to three-dimensional images. Beyond synthesis, BUSGen \cite{yu2026busgen} shows how generative pretraining and adaptation can support breast cancer screening, diagnosis and prognosis. These advances establish a broader role for medical generators, but repeated denoising increases the cost of producing large image collections. They motivate an alternative that generates images efficiently, improves image quality as model size increases and supports subsequent medical image analysis.

Autoregressive generation offers a different way to learn and sample image distributions. In medical imaging, Tudosiu et al. \cite{tudosiu2024realistic} used sequential token prediction to generate brain images. Visual autoregressive modelling (VAR) \cite{tian2024visual} instead predicts progressively finer token maps, generating all tokens within a scale in parallel. This next-scale formulation improves quality and speed on natural images, with quality improving as model size increases. FedGAT \cite{nezhad2025fedgat} applies next-scale prediction to federated magnetic resonance imaging (MRI) reconstruction, while CVARPath \cite{liu2025cvarpath} restores pathological images. These applications focus on particular anatomical or task settings. Broader synthesis across computed tomography (CT) and MRI requires learning anatomical and appearance variation while retaining medically informative features. Extending next-scale autoregression to this setting offers a route to a medical foundation model that combines efficient generation with control over anatomy and supports subsequent image analysis.

\begingroup
\clubpenalty=10000
Here we present MedVAR, a generative foundation model that learns across six anatomical regions while allowing users to specify anatomical content and spatial structure. To our knowledge, it is the first next-scale autoregressive generator for CT and MRI across anatomical regions. We trained a multiscale tokenizer on medical images and introduced hierarchical semantic conditioning and structural attention routing. The semantic pathway combines region and organ information with modality, while the structural pathway uses masks and boundaries to guide spatial layout. Across 40 datasets, including data from seven internal clinical centres, MedVAR achieves lower Fréchet inception distance and shorter sampling latency than the evaluated diffusion baselines at comparable parameter counts. Generation quality improves as model size increases. Its generated images also improve multicentre hepatocellular carcinoma segmentation and support missing-slice reconstruction, linking the generative model to downstream analysis. Privacy analyses further distinguish information exposed through model likelihoods from similarity between generated and training images.
\par
\endgroup

%% file: 2_results_recognition.tex
\subsection{MedVAR improves generation quality and speed over diffusion baselines}

With semantic conditions alone, MedVAR combined lower Fr\'echet inception distance (FID) with faster sampling than the diffusion baselines at comparable parameter counts. We evaluated one generator across six anatomical regions using 482,815 images from 40 datasets and cohorts: 33 public collections and seven internal clinical centres. The collection comprised 366,888 CT and 115,927 MRI images; 438,905 images were used for training and 43,910 for reference evaluation (Fig.~1a, b and Supplementary Table~1). The internal hepatocellular carcinoma (HCC) cohort contributed 1,082 T2-weighted MRI volumes, with 60--417 volumes per centre and visible variation in tumour appearance (Fig.~1d). MedVAR generated images through progressively finer token maps, guided by semantic labels and optional masks and boundaries (Fig.~1c).

\begin{figure*}[htbp]
    \centering
    \includegraphics[width=\linewidth,height=0.74\textheight,keepaspectratio]{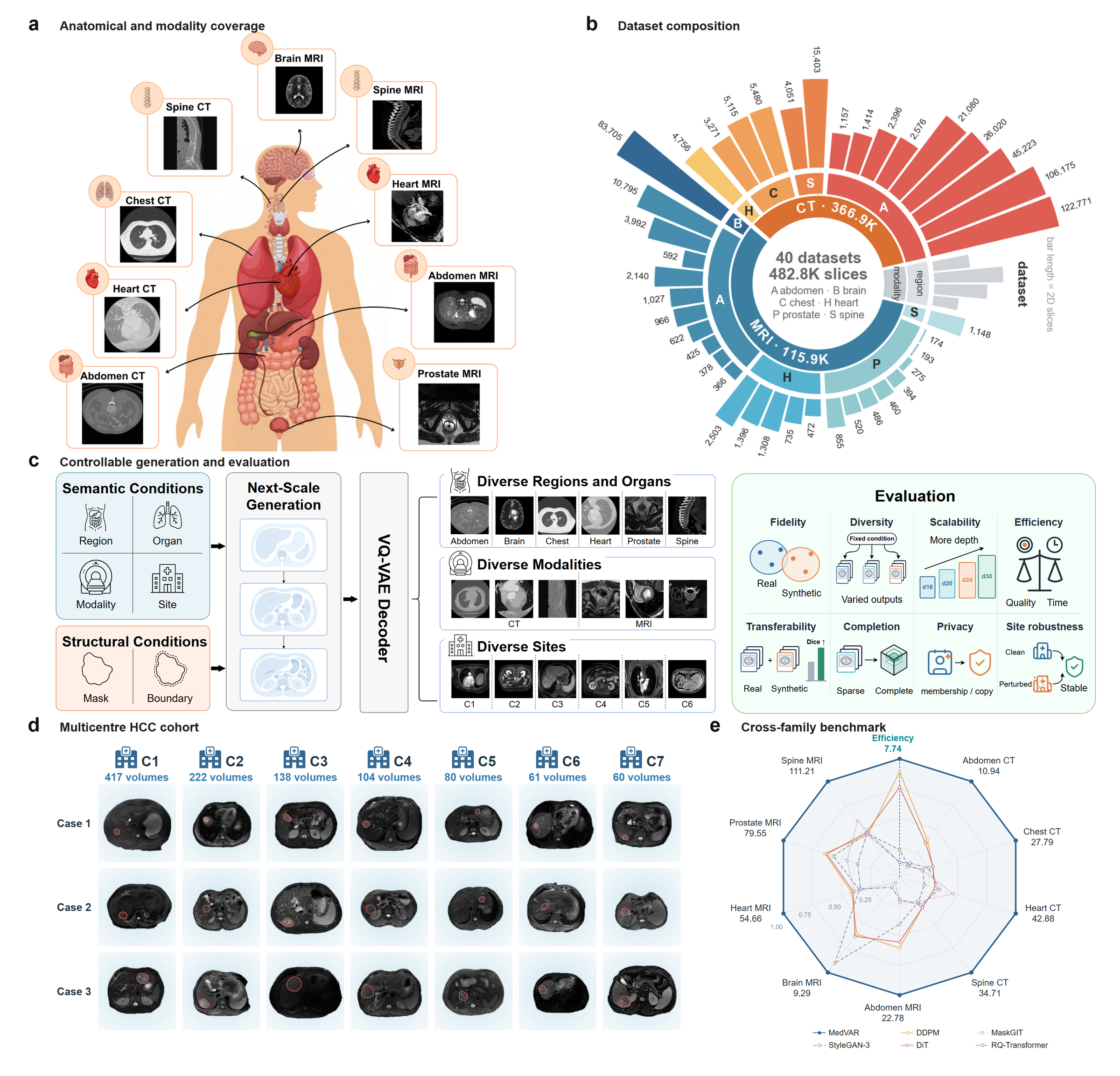}
    \caption{\figureonecaption}
    \label{fig:data-composition}
\end{figure*}

At approximately 330 million parameters, depth-16 MedVAR achieved an FID of 11.01 in 0.110~s per image. Denoising diffusion probabilistic models (DDPM) \cite{ho2020ddpm} and diffusion transformers (DiT) \cite{peebles2023scalable} were trained on the same images and data source labels. Their 330-million-parameter configurations, DDPM-S and DiT-S, achieved FIDs of 11.63 and 20.85 in 1.55 and 1.39~s, respectively, using 100 denoising steps. At approximately 600 million parameters, depth-20 MedVAR achieved an FID of 10.43 in 0.131~s. The corresponding DDPM-L and DiT-L models achieved 10.56 in 1.55~s and 13.44 in 2.43~s. Thus, MedVAR achieved lower FID with approximately 12--19 times faster sampling across the two model sizes (Fig.~2d).

Reducing DDPM and DiT from 100 to 10 denoising steps shortened latency but increased FID (Fig.~2d). MedVAR also retained its efficiency in a batch benchmark excluding file writing. At batch size 64, MedVAR combined lower FID with higher throughput than similarly sized diffusion baselines (Supplementary Fig.~1a and Supplementary Table~2). Increasing batch size raised throughput and lowered time per image at every MedVAR depth (Supplementary Fig.~1b, c and Supplementary Table~3). At batch size 64, throughput ranged from 50.35 images per second at depth 30 to 131.98 at depth 16. These measurements show that the sampling advantage extended beyond single-image generation.

\subsection{Image quality improves with increasing MedVAR size}

Increasing MedVAR depth improved image quality with both semantic and structural conditions: FID decreased from 10.87 to 10.15, 9.78 and 9.31 at depths 16, 20, 24 and 30, respectively. With semantic conditions alone, the same checkpoints yielded FIDs of 11.01, 10.43, 10.16 and 10.03, with latencies of 0.110--0.189~s per image. This configuration supports comparison with generators without structural inputs. Examples across depths show the changes in anatomical detail (Fig.~2b, d).

The quality advantage was also present within individual imaging settings. With semantic conditions alone, depth-30 MedVAR ranked first by FID in all nine groups defined by anatomical region and modality among methods with available results (Fig.~2c and Supplementary Table~4). Against the five baselines trained for this study, the same configuration also had the highest normalized scores in the overview of quality and efficiency (Fig.~1e). The wider comparison included medical generators and general text-to-image systems; the latter were evaluated without medical training as contextual references. Matched chest CT, heart MRI and prostate MRI examples showed recognizable anatomy from MedVAR, whereas several comparator images contained missing or distorted structures (Fig.~2a).

\looseness=-1
We assessed additional aspects of image quality with kernel inception distance (KID), RadImageNet FID (RadFID) and CLIP maximum mean discrepancy (CMMD). MedVAR achieved the lowest KID and CMMD among the evaluated generator families (Fig.~2e). Adding structural conditions to depth-30 MedVAR reduced FID from 10.03 to 9.31 and CMMD from 0.165 to 0.152. KID remained 0.003 and RadFID remained 0.03 at the displayed precision.

The generated images also retained information about the requested anatomy and modality. A classifier trained on real images measured condition recognizability, while classification accuracy score (CAS) \cite{ravuri2019classification} measured performance after training on generated images and testing on real images. Across depths, recognizability ranged from 0.948 to 0.959 and CAS from 0.952 to 0.981 with and without structural conditions (Fig.~2f and Supplementary Table~5). The highest corresponding values among the other models were 0.609 and 0.661, both obtained by MaskGIT \cite{chang2022maskgit}. The task comprised nine classes, with chance balanced accuracy of $1/9$. The classifier trained on real images reached 0.9999 on held-out real images (95\% confidence interval, 0.9998--0.9999), providing a reference for recognizability.

MedVAR generated varied images within each condition. Its multiscale structural similarity was lower and its perceptual distance was higher than those of StyleGAN3 \cite{karras2021alias}, with variation also exceeding the held-out real reference (Fig.~2g). Changing only the random seed produced distinct abdomen CT, spine CT and brain MRI images while preserving recognizable anatomy (Supplementary Fig.~2). Together with recognizability and CAS, these findings show that low FID was accompanied by both semantic consistency and variation within conditions. We next examined the representation and conditioning choices that supported these properties.

\begin{figure*}[htbp]
    \centering
    \includegraphics[width=\linewidth,height=0.925\textheight,keepaspectratio,trim=0 194bp 0 0,clip]{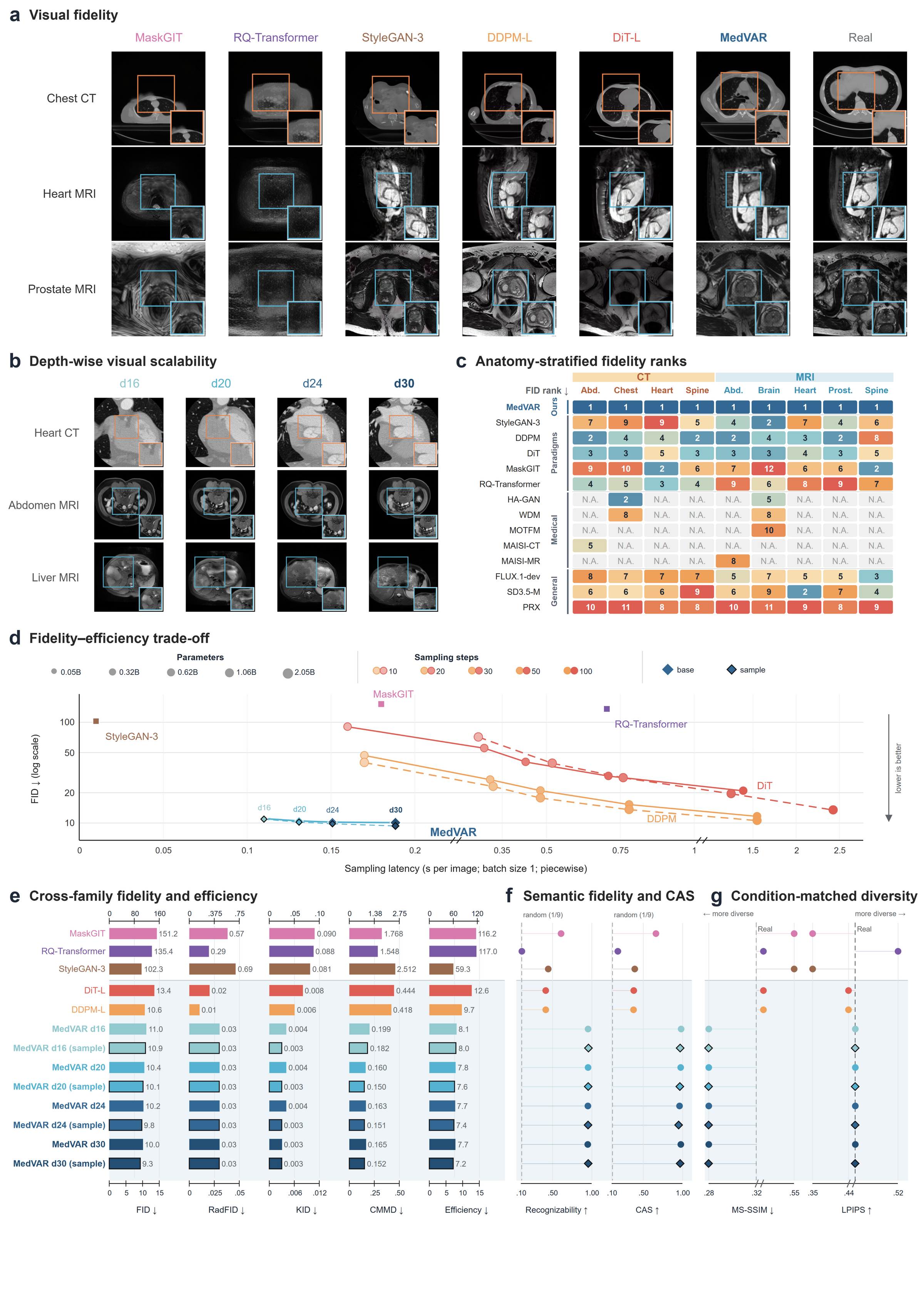}
    \caption{\figuretwocaption}
    \label{fig:cross-family-evaluation}
\end{figure*}

\subsection{Medical tokenization and conditioning enable explicit control}

Training the tokenizer on medical images improved reconstruction fidelity and codebook utilization (Fig.~3a). Across 43,910 held-out images, reconstruction peak signal-to-noise ratio (PSNR) increased from $29.88\pm2.31$~dB with the ImageNet tokenizer to $32.70\pm2.91$~dB with the medical tokenizer. Values are means $\pm$ sample standard deviations. The mean paired gain was 2.81~dB (95\% confidence interval from bootstrap resampling by dataset, 2.10--3.37~dB). Gains were positive for 92.7\% of images and on average in all 40 datasets. The mean gain was 2.70~dB in CT and 3.18~dB in MRI (Supplementary Table~6). Codebook utilization increased from 97.0\% to 99.2\% in the fixed evaluation subset. The medical tokenizer also supported recognizable CT and MRI generation where the ImageNet tokenizer produced incoherent images.

Semantic conditions specified image content, while masks and boundaries provided spatial control. Adding anatomical and modality labels changed the generated anatomy and acquisition appearance; adding structural inputs further constrained spatial arrangement (Fig.~3b). With the data source label alone, FID was 16.59. Adding organ and modality labels reduced it to 12.02; the full model with semantic conditions alone reached 11.01. Supplying structural conditions further reduced FID to 10.87. This setting gave the lowest KID, CMMD, weighted FID, macro FID and tail FID, with a joint lowest RadFID (Fig.~3c and Supplementary Table~7). With the seed and other inputs fixed, translating or dilating the mask changed the location or extent of the corresponding structure (Supplementary Fig.~3).

The scale at which structural information entered attention affected the balance between overall and group results. We compared early, intermediate and late routing, and removed the routing gates or auxiliary structure head in separate ablations (Fig.~3d). Intermediate routing at token-map sizes $5^2$, $6^2$, $8^2$ and $10^2$ gave the lowest FID with semantic conditions alone (11.01). With structural inputs, it gave the lowest macro FID (62.13) and tail FID (80.75). Late routing gave lower overall FID (10.65) and weighted FID (21.98), but higher macro and tail FID. Removing gates or the auxiliary head did not improve all measures. We selected intermediate routing to balance overall image quality with performance across groups.

The structural weight and sampling parameters provided further control over image quality. A structural weight of $\alpha=0.25$ gave the lowest or joint lowest values across all seven measures, including FID 10.87 (Fig.~3e). Across classifier-free guidance (CFG) values of 3.5--5.5, FID varied by 0.13; RadFID and CMMD were unchanged at the displayed precision. CFG 4.5 reduced KID from 0.004 to 0.003 and was retained for subsequent generation. With semantic conditions alone, top-$k$ 1,300 and top-$p$ 0.98 gave the lowest overall FID (11.01), while 1,100 and 0.97 gave slightly lower group summaries (Fig.~3f).

\begin{figure*}[htbp]
    \centering
    \includegraphics[width=\linewidth,height=0.9\textheight,keepaspectratio]{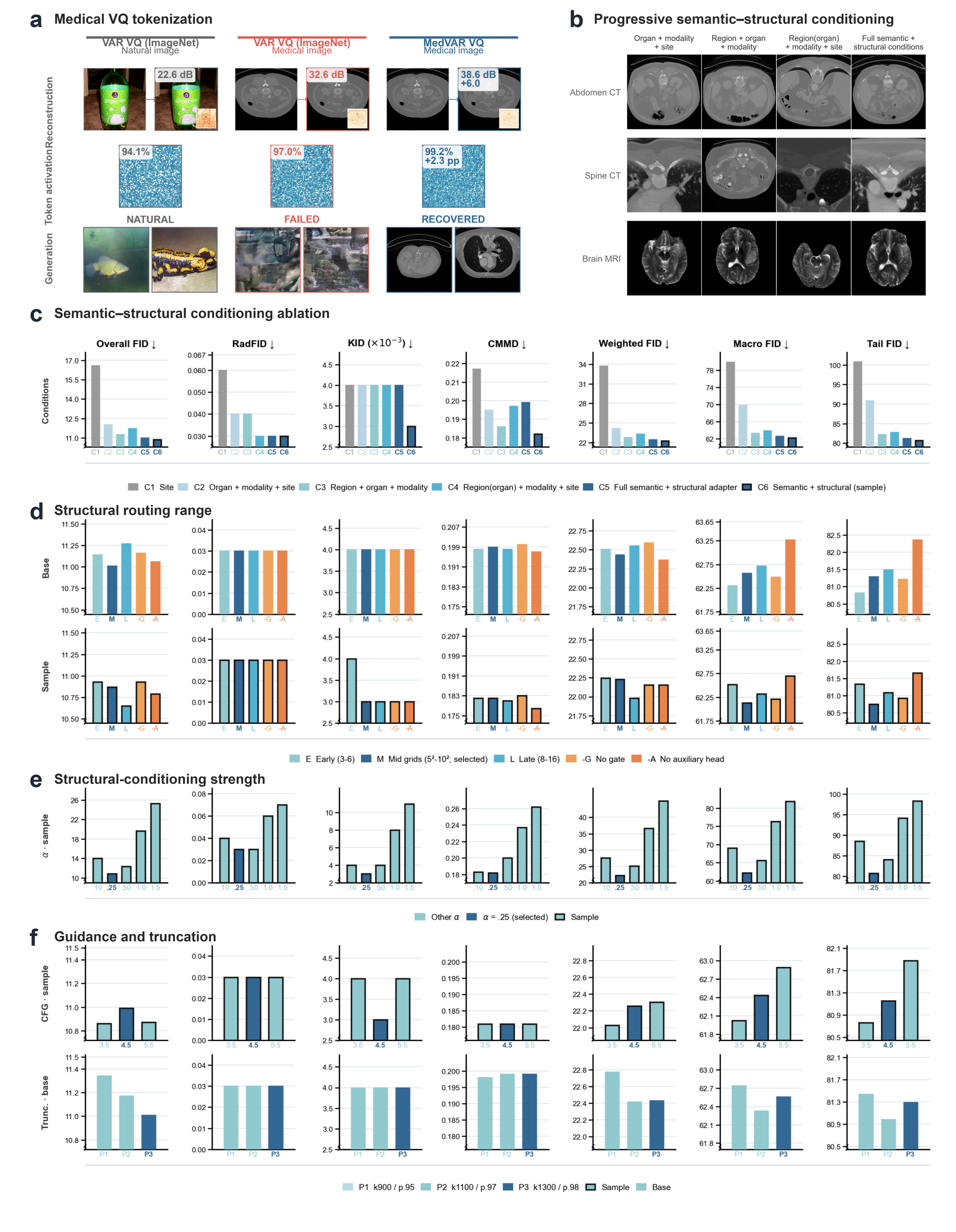}
    \caption{\figurethreecaption}
    \label{fig:ablation}
\end{figure*}

\subsection{Pretraining on generated images supports multicentre HCC segmentation}

We next tested whether the generated images supported learning from patient data. For HCC segmentation, a U-Net was pretrained on real images, MedVAR images generated with centre labels or unconditional MedVAR images (Fig.~4a). The three groups used identical image counts and schedules, followed by the same pretraining on real images and supervised fine-tuning. Fine-tuning used 40\% of training cases from each centre. The labelled subsets, held-out volumes and training seeds were matched across groups; generated lesion masks were not used for supervision (Methods).

Pretraining on MedVAR images improved Dice relative to pretraining on real images. Macro-averaged volume Dice across centres increased from $0.6151\pm0.0057$ with real images to $0.6324\pm0.0213$ with images generated using centre labels. These values are means $\pm$ sample standard deviations across three training seeds. The paired mean gain was 0.0173 and was positive for all three seeds. Mean gains were positive at all seven centres and largest at C5 (Fig.~4b and Supplementary Table~8). Unconditional MedVAR images also improved Dice ($0.6271\pm0.0057$; paired mean gain, 0.0120). Both generated image sets therefore supported representation learning, with the larger mean gain obtained using centre labels.

The generated images retained some of the appearance associated with acquisition centres. In the displayed examples, images generated using centre labels retained abdominal MRI appearance, while two unconditional examples departed from this setting (Fig.~4c). Principal component analysis (PCA) and t-distributed stochastic neighbour embedding (t-SNE) showed broad overlap among centres (Fig.~4d). A classifier identified the requested centre in 36.0\% of generated images on average, compared with 14.3\% chance. Accuracy ranged from 12\% to 73\% across centres, and the feature discrepancy between real and generated images also varied by centre (Fig.~4e). Centre labels thus retained detectable acquisition differences without producing isolated clusters in these visualizations.

\subsection{MedVAR images support reconstruction of missing slices}

MedVAR images also served as an additional input for reconstructing missing slices (Fig.~4a, f). The frozen generator used two observed neighbouring slices to produce a deterministic image, which a separate reconstruction network combined with those observations. The target slice was withheld from all inputs and used only for supervision and evaluation. In the displayed CT example, learned reconstruction restored structures blurred by linear interpolation; difference maps show local changes between the U-Net and the reconstruction system using MedVAR (Fig.~4f). Across 40 held-out CT volumes, reconstruction using MedVAR achieved $37.8326\pm0.0223$~dB PSNR across three training seeds. The independently trained U-Net achieved 37.7608~dB and linear interpolation 35.9403~dB. The mean gain over the U-Net was 0.0718~dB (95\% bootstrap confidence interval, 0.0357--0.1167). Across 40 held-out MRI volumes, the corresponding values were $23.3419\pm0.0348$, 23.2814 and 21.9343~dB, with a mean gain of 0.0605~dB (0.0162--0.1034). Gains were positive for all three seeds in both modalities (Supplementary Table~9).

The reconstructed CT stacks largely preserved volumetric segmentation performance. We applied the same frozen three-dimensional nnU-Net \cite{isensee2021nnunet} to reconstructed and fully sampled volumes (Fig.~4g). Across 36 held-out volumes with a fixed subset of interior slices removed, mean Dice was $0.60544\pm0.00037$ after reconstruction using MedVAR and 0.60721 for fully sampled CT. Of the 36 reconstructed volumes, 33 remained within 0.01 Dice of their fully sampled counterparts.

\begin{figure*}[htbp]
    \centering
    \includegraphics[width=\linewidth,height=0.74\textheight,keepaspectratio]{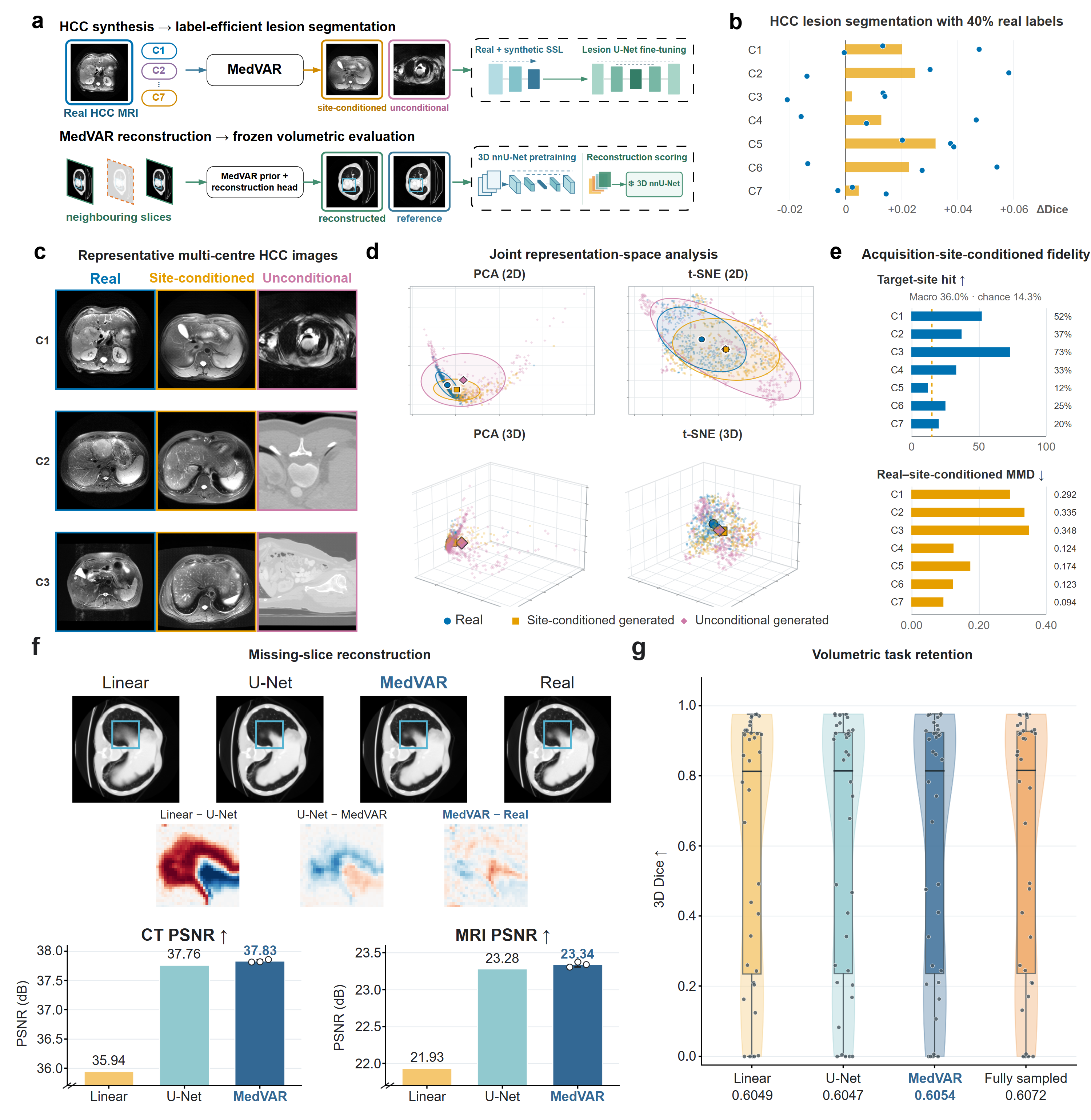}
    \caption{\figurefourcaption}
    \label{fig:downstream-utility}
\end{figure*}

\subsection{Privacy analyses distinguish model access from generated image release}

Access to model likelihoods revealed a modest signal of training membership for the depth-16 checkpoint. The white-box analysis \cite{shokri2017membership,kowalczuk2025privacy} included 43,828 matched slice pairs, with separate cases used for calibration and testing (Fig.~5a, b and Supplementary Table~10). The published Min-K 40\% score \cite{shi2024detecting} achieved a receiver operating characteristic area under the curve (ROC AUC) of 0.5949 (95\% bootstrap confidence interval, 0.5891--0.6007). Its true-positive rate (TPR) was 3.48\% at a 1\% false-positive rate (FPR). A multiscale score gave the largest separation: ROC AUC 0.6067 (0.6005--0.6131) and TPR 3.54\% (3.12--3.95\%) at the same operating point. Controls using metadata or image intensities remained close to chance, indicating that these variables alone did not explain the membership signal.

\looseness=-1
The analysis of generated images gave a different result. Nearest-neighbour copy detection \cite{dar2026memorization} tested whether generated images were closer to training images than to held-out images. The analysis included 4,705 paired cases and 86,922 candidate images, three independently trained encoders and three generated set sizes (Fig.~5c, d and Supplementary Table~11). Across set sizes, mean ROC AUC ranged from 0.5018 to 0.5024 and TPR at 1\% FPR from 1.10\% to 1.12\%. These values remained close to chance under the tested attack. The displayed examples included close neighbours from both training and test sets (Fig.~5d).

Alongside privacy, we assessed generation reliability by testing whether data source labels helped preserve quality when the training collection contained unrelated images. A $2\times2$ experiment varied label use and the presence of simulated contamination while keeping other settings fixed (Fig.~5e and Supplementary Table~12). Providing the label lowered FID against each source's reference images in all four datasets, both with and without contamination. The interaction between label use and contamination was positive for AMOS, BTCV and WORD and negative for AbdomenCT-1K. The labels therefore improved fidelity to the requested source, but their response to contamination varied across datasets.

\begin{figure*}[htbp]
    \centering
    \includegraphics[width=\linewidth,height=0.74\textheight,keepaspectratio]{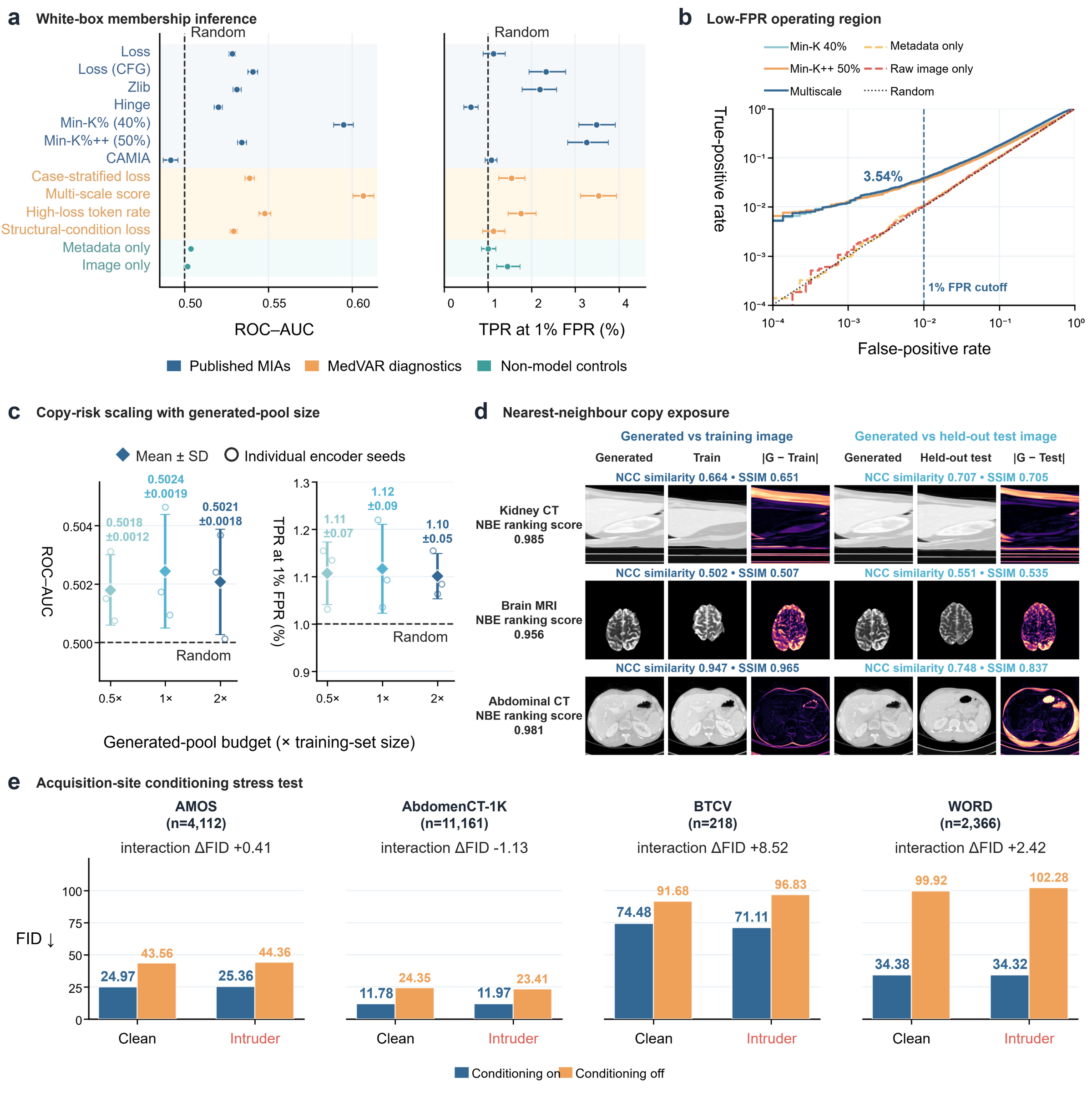}
    \caption{\figurefivecaption}
    \label{fig:privacy-reliability}
\end{figure*}

%% file: 8_discussion.tex
MedVAR combines high image quality with efficient sampling across anatomical regions in CT and MRI. With semantic conditions alone, it achieved lower FID and shorter latency than DDPM and DiT at comparable parameter counts. With both semantic and structural conditions, increasing model size reduced FID from 10.87 to 9.31. This extends medical autoregression beyond brain synthesis \cite{tudosiu2024realistic}, federated MRI reconstruction with FedGAT \cite{nezhad2025fedgat} and pathological restoration with CVARPath \cite{liu2025cvarpath}, providing controllable images for subsequent learning and reconstruction.

\looseness=-1
The efficiency advantage is consistent with how MedVAR organizes prediction. Whereas diffusion repeatedly denoises an image representation, MedVAR predicts progressively finer token maps, with all tokens at each scale generated in parallel. In the reported comparison with 100-step diffusion, this meant fewer sequential network evaluations. VAR \cite{tian2024visual} established this approach for natural images; MedVAR retains its sampling advantage after medical adaptation. The batch measurements further show that this advantage extends to generating image collections.

Quality improved with increasing model size both with and without structural inputs. Greater capacity and longer training may support more effective use of semantic labels and spatial guidance. Coarser token maps provide context for finer predictions, allowing larger transformers to better model anatomical structure and imaging appearance. This offers a possible explanation for the trend, alongside the improvements with model size reported for VAR \cite{tian2024visual}.

Medical tokenization and conditioning address requirements beyond the original class-conditioned VAR model. The medical tokenizer reduced reconstruction error and supported recognizable generation where the ImageNet tokenizer failed, suggesting better preservation of tissue contrast and anatomical detail. Hierarchical semantic conditioning combines region and organ information before integrating modality, while structural attention routing uses masks and boundaries to guide spatial refinement. These complementary pathways may help vary appearance while preserving the requested anatomy, as supported by the condition ablations and mask transformations. Routing structural information at intermediate scales may guide refinement without dominating the finest details, consistent with the macro and tail FID results.

Image quality requires complementary measurements. With semantic conditions alone, MedVAR had the lowest FID across nine anatomical and modality groups and the lowest overall KID and CMMD; structural inputs further improved FID and CMMD. RadFID favoured some diffusion baselines, reflecting the sensitivity of distributional metrics to feature representations \cite{woodland2024_fid_med,jayasumana2024rethinkingfidbetterevaluation}. Recognizability and CAS test whether generated images retain the requested content, while diversity measures assess variation within conditions. Together, these measures characterize distributional quality, semantic consistency and diversity.

The downstream findings suggest that generated images retain information useful for learning and reconstruction. MINIM \cite{wang2025minim} uses synthetic images for clinical analysis, while BUSGen \cite{yu2026busgen} also demonstrates how adapting a generator can support diagnosis and prognosis. For segmentation, synthetic pretraining may expose the encoder to alternative examples of anatomy and acquisition appearance before supervised fine-tuning. Gains over pretraining on equally sized real image sets show that the generated set provided useful training information under the matched schedule. For reconstruction, the MedVAR image may provide an anatomical estimate that complements the local information in neighbouring slices. The reconstruction network combines this estimate with observed slices. Positive PSNR gains across CT and MRI, together with Dice close to the fully sampled CT reference, support the utility of this reconstruction system.

For the depth-16 model, the different privacy results can be understood in terms of what each attack observes. A likelihood-based attack uses probabilities assigned to candidate images \cite{kowalczuk2025privacy}, whereas copy detection compares generated outputs with training and reference images \cite{dar2026memorization}. Small probability differences can expose membership without requiring reproduction of a training image. A membership signal in token likelihoods can therefore coexist with near-chance copy detection. This distinction motivates separate assessments of model access and image release. Beyond privacy, data source labels may help distinguish the requested image distribution from unrelated training images. Their lower FID in both clean and contaminated settings supports this role, although the response to contamination varied across datasets.

The current evidence comes from two-dimensional CT and MRI synthesis and retrospective downstream studies. Diffusion comparisons matched parameter counts near 330 million and 600 million; larger-scale analyses varied MedVAR size and training duration. Future work could adapt the pretrained generator directly for recognition and segmentation, extend it to volumes, validate downstream models in independent cohorts and assess image validity with clinicians.

Together, these findings establish next-scale autoregression as an efficient approach to high-quality medical image generation across anatomical regions. Medical tokenization, hierarchical semantic conditioning and structural attention routing adapt this framework to medical imaging. MedVAR achieves lower FID and faster sampling than the evaluated diffusion baselines, with controlled outputs that support downstream analysis.

%% file: 5_method.tex
\subsection{Study design}

We compared quality and sampling speed between MedVAR with semantic conditions alone and DDPM and DiT at two approximately matched parameter counts. The complete semantic and structural conditioning configuration was evaluated across four MedVAR sizes, while diffusion steps varied at fixed model sizes. Ablations tested medical tokenization, conditioning and sampling parameters. With the generator fixed, generated images supported pretraining for multicentre hepatocellular carcinoma (HCC) lesion segmentation with limited labelled data and reconstruction of missing slices. White-box membership inference and nearest-neighbour copy detection assessed model access and image release, respectively. A separate experiment tested data source labels under simulated contamination.

\subsection{Data sources and composition}

The full image collection contains 482,815 two-dimensional images from 40 public and internal datasets. It comprises 366,888 CT and 115,927 MRI images spanning abdomen, brain, chest, heart, prostate and spine. Training used 438,905 images (333,526 CT and 105,379 MRI). The held-out reference set contained 43,910 images (33,362 CT and 10,548 MRI) for the main generation analyses. Public data include established abdominal, thoracic, neurological, cardiac, prostate and musculoskeletal benchmarks and were used under their original access and licensing terms. Dataset composition and public access information are provided in Supplementary Table~1. Each sample is associated, where available, with an anatomical region, one or more organ labels, imaging modality, data source and a foreground segmentation mask. For public collections, the source label identifies the dataset; for the internal HCC cohort, it identifies the anonymized acquisition centre.

\looseness=-1
The internal HCC cohort contains 1,082 preprocessed T2-weighted MRI volumes from seven anonymized centres. The source acquisitions cover 1.5- and 3-T scanners from three anonymized vendors. Lesion masks were available for all volumes. For the HCC utility analysis, 5,382 lesion-positive processed slices were assigned by patient to 758 training, 108 validation and 216 test volumes using a centre-stratified split with seed 9101. Only anonymized centre identifiers and aggregated acquisition characteristics are used by the model or reported in the study.

\subsection{Image and mask preprocessing}

We converted each source volume into three-channel slices with intensities in $[0,1]$. Intensity normalization was performed during preparation of each dataset. No additional common CT window or MRI percentile normalization was applied. During transformer training, the shorter image dimension was resized to 288 pixels followed by a random $256\times256$ crop. Validation and evaluation used a centred crop. Random horizontal flipping was not used. Images and masks were transformed with the same geometry, with nearest-neighbour interpolation for masks.

Available class masks were converted to a binary foreground union. A structural boundary map was computed as the morphological gradient of the foreground mask using a $3\times3$ neighbourhood and paired with the mask as a two-channel structural condition. Samples without valid masks were retained for training with semantic conditions, using a zero-valued structural input and a mask validity indicator. This prevented missing annotations from being treated as background evidence.

\subsection{Medical multiscale tokenizer}

MedVAR uses a residual vector-quantized autoencoder to convert an image $x$ into a hierarchy of discrete token maps $\{r_k\}_{k=1}^{K}$ (Fig.~1c). The tokenizer follows the multiscale residual formulation of VAR \cite{tian2024visual} and was trained on the medical collection. For $256\times256$ images, the ten token-map resolutions are
\begin{equation}
(1,2,3,4,5,6,8,10,13,16),
\end{equation}
which together contain 680 autoregressive token positions. The shared codebook contains 4,096 entries with 32-dimensional embeddings. At each scale, the aligned residual feature is quantized, decoded as a refinement and subtracted before the next finer-scale quantization. The tokenizer loss combines L1 reconstruction, L2 reconstruction and perceptual terms with weights 0.2, 1.0 and 0.5, respectively, and uses a vector-quantization commitment coefficient of 0.25. Once trained, the tokenizer is frozen for all autoregressive experiments.

To compare tokenization for natural and medical images, we applied the frozen ImageNet-trained VAR tokenizer and the medically trained tokenizer to the identical set of 43,910 held-out medical images. We calculated per-image reconstruction PSNR and summarized paired differences overall, by modality and by dataset. Uncertainty in the overall paired gain was estimated with 20,000 bootstrap draws that resampled the 40 datasets as clusters and retained all images within each selected dataset. A fixed subset of 400 medical and 100 natural images measured codebook utilization. It also provided examples of autoregressive generation with matched seeds. 

\subsection{Next-scale autoregressive modelling}

MedVAR factorizes the conditional probability of coarse-to-fine token maps as illustrated in Fig.~6a:
\begin{equation}
p(r_1,\ldots,r_K\mid c,d)=\prod_{k=1}^{K}p_{\theta}\left(r_k\mid r_{<k},c,d\right),
\end{equation}
\looseness=-1
where $c$ denotes the low-dimensional semantic conditions, including the separately routed embedding of the source label, and $d$ denotes the optional spatial structural condition. During training, all scales are supplied by teacher forcing and the transformer minimizes cross-entropy over the 680 image-token positions. During inference, each scale is sampled once, appended to the token pyramid and decoded by the frozen tokenizer after the final scale.

MedVAR was evaluated at transformer depths of 16, 20, 24 and 30 blocks. Their transformer parameter counts are 322.2, 619.8, 1,060.9 and 2,053.4 million. The depth-16 and depth-20 models provided approximate parameter matches to the 330-million-parameter and 600-million-parameter DDPM and DiT baselines, respectively. The depth-24 and depth-30 models extended the analysis of increasing size within MedVAR. MedVAR models were optimized with fused AdamW ($\beta_1=0.9$, $\beta_2=0.95$) in bfloat16 using an effective batch size of 768 and a peak learning rate of $1.8\times10^{-4}$. Gradient clipping was set to 2, label smoothing to 0.05 and weight decay increased from 0.05 to 0.08. The learning rate was warmed up for 2\% of training, held at its peak for 5\% of the post-warm-up schedule and then linearly decayed to 1\% of the peak. The depth-16 and depth-20 models were trained for 240 epochs, the depth-24 model for 300 epochs and the depth-30 model for 350 epochs. Model training and the main latency measurements used NVIDIA H200 GPUs. Latency was measured at batch size 1 with matched hardware and implementation settings within the reported comparison. A separate batch inference benchmark used one fixed abdomen CT condition with semantic conditions alone and excluded file writing while retaining VQ decoding. For each MedVAR depth, batch sizes 1, 2, 4, 8, 16, 32 and 64 were measured after two warm-up iterations; medians and interquartile ranges were calculated from ten repetitions. The comparison across generator families used the corresponding measurements excluding file writing at batch sizes 1 and 64.

\subsection{Hierarchical semantic conditioning}

Semantic conditioning combines anatomical and modality information with information about the data source (Fig.~6b). Anatomical region and modality use learned embeddings. Organ labels are represented by a multi-hot vector; the embeddings of active organs are averaged and projected within the region representation. Region, organ and modality embeddings are fused, whereas the source label is encoded separately and contributes a scaled embedding to adaptive layer normalization, with coefficient 0.10. Together, these inputs define semantic conditioning. Semantic inputs are low-dimensional categorical or multi-hot variables, whereas the structural input described below is a spatially dense mask--boundary pyramid. In addition to the global conditioning vector, a scale-aware token bias was added with coefficient 0.10. For scale $k\in\{1,\ldots,K\}$, its normalized position and unnormalized region, organ and modality weights were
\begin{align}
t_k&=\frac{k-1}{K-1},\\
\widetilde{w}_{\mathrm{reg},k}&=(1-t_k)^{1.5}+0.05,\\
\widetilde{w}_{\mathrm{org},k}&=\max\{0,1-|2t_k-1|\}+0.05,\\
\widetilde{w}_{\mathrm{mod},k}&=t_k^{1.5}+0.05.
\end{align}
The weights were normalized before the corresponding embeddings were combined,
\begin{equation}
w_{a,k}=\frac{\widetilde{w}_{a,k}}
{\widetilde{w}_{\mathrm{reg},k}+\widetilde{w}_{\mathrm{org},k}+\widetilde{w}_{\mathrm{mod},k}},
\qquad a\in\{\mathrm{reg},\mathrm{org},\mathrm{mod}\}.
\end{equation}
During training, the fused anatomy--modality component was replaced by the null condition with probability 0.10. Data source and valid structural conditions were independently dropped with probabilities 0.60 and 0.20, respectively. Conditional dropout enabled classifier-free guidance at inference.

Condition ablations compared six configurations. The first three used data source alone; organ, modality and data source; or region, organ and modality. The fourth added hierarchical region--organ conditioning to modality and data source. The fifth used all semantic conditions with the structural pathway present but no structural input. The sixth additionally supplied structural conditions. We report overall FID together with weighted, macro and tail FID across anatomy--modality groups. 

\subsection{Structural attention routing}

The structural condition comprises a binary foreground mask $M$, its boundary $B$ and a validity bit $v$ (Fig.~6c). Given an available segmentation label map $S$, the foreground union and its $3\times3$ morphological gradient were
\begin{align}
M&=\mathbf{1}[S>0],\\
B&=\operatorname{dilate}_{3\times3}(M)-\operatorname{erode}_{3\times3}(M).
\end{align}
Let $\mathcal{P}_{k}$ denote adaptive average pooling to the token-map resolution at scale $k$. For token $i$, the pooled mask and boundary values and their projected structural feature were
\begin{equation}
m_{k,i}=[\mathcal{P}_{k}(M)]_i,\qquad
b_{k,i}=[\mathcal{P}_{k}(B)]_i,\qquad
\mathbf d_{k,i}=f_{\phi}([m_{k,i},b_{k,i}]),
\end{equation}
where $f_{\phi}$ is a layer-normalized multilayer perceptron that maps the two continuous channels to the transformer hidden dimension. At each scale, these features modify only the query and key projections of self-attention:
\begin{align}
\Delta\mathbf q_{k,i}
&=\alpha v\rho_k g_k^{(\mathrm Q)}(0.10m_{k,i}+0.90b_{k,i})\mathbf d_{k,i},\\
\Delta\mathbf k_{k,i}
&=\alpha v\rho_k g_k^{(\mathrm K)}(0.55m_{k,i}+0.45b_{k,i})\mathbf d_{k,i},\\
\mathbf q'_{k,i}&=\mathbf q_{k,i}+\Delta\mathbf q_{k,i},\qquad
\mathbf k'_{k,i}=\mathbf k_{k,i}+\Delta\mathbf k_{k,i}.
\end{align}
Here, $g_k^{(\mathrm Q)}$ and $g_k^{(\mathrm K)}$ are learned gates that include a scale gate initialized to zero. The routing multiplier is $\rho_k=1$ for token-map sizes $5^2$, $6^2$, $8^2$ and $10^2$, and $\rho_k=0.05$ otherwise. The structural condition weight is $\alpha=0.25$ in the evaluated setting. The validity bit sets the update to zero when a structural condition is absent. Mask and boundary features are not added to the residual stream, used as prefix tokens or predicted as part of the autoregressive target; value and output projections are unchanged.

For the auxiliary structure-prediction task, pooled values were converted to four class targets,
\begin{align}
y_{k,i}&=\mathbf{1}[m_{k,i}\geq0.50]+2\mathbf{1}[b_{k,i}\geq0.25],\\
\mathcal{L}&=\mathcal{L}_{\mathrm{AR}}+0.10\mathcal{L}_{\mathrm{struct}}.
\end{align}
The auxiliary head was applied only at token-map sizes $5^2$, $6^2$, $8^2$ and $10^2$. It comprised layer normalization, a linear projection from $C$ to $C/4$, SiLU activation and a four-class linear projection. Boundary observations were weighted by 2 in the cross-entropy term. The head regularized the structural representation during training and was removed at inference.

Two generation settings were evaluated with the same trained model. The setting labelled ``base'' in the figures used semantic conditions alone. The setting labelled ``sample'' additionally supplied an anatomical mask and its boundary pyramid to the attention router. Both settings used classifier-free guidance (CFG) 4.5, top-$k$ 1,300 and top-$p$ 0.98 in the main generation analysis; the structural input was weighted by $\alpha=0.25$. Qualitative structural responses were tested by translating the mask rightward, translating it downward or dilating it. Semantic conditions, sampling parameters and the seed were fixed within each example. The displayed contours denote the supplied masks rather than predicted segmentations.

\begin{figure*}[htbp]
    \centering
    \includegraphics[width=0.9\linewidth]{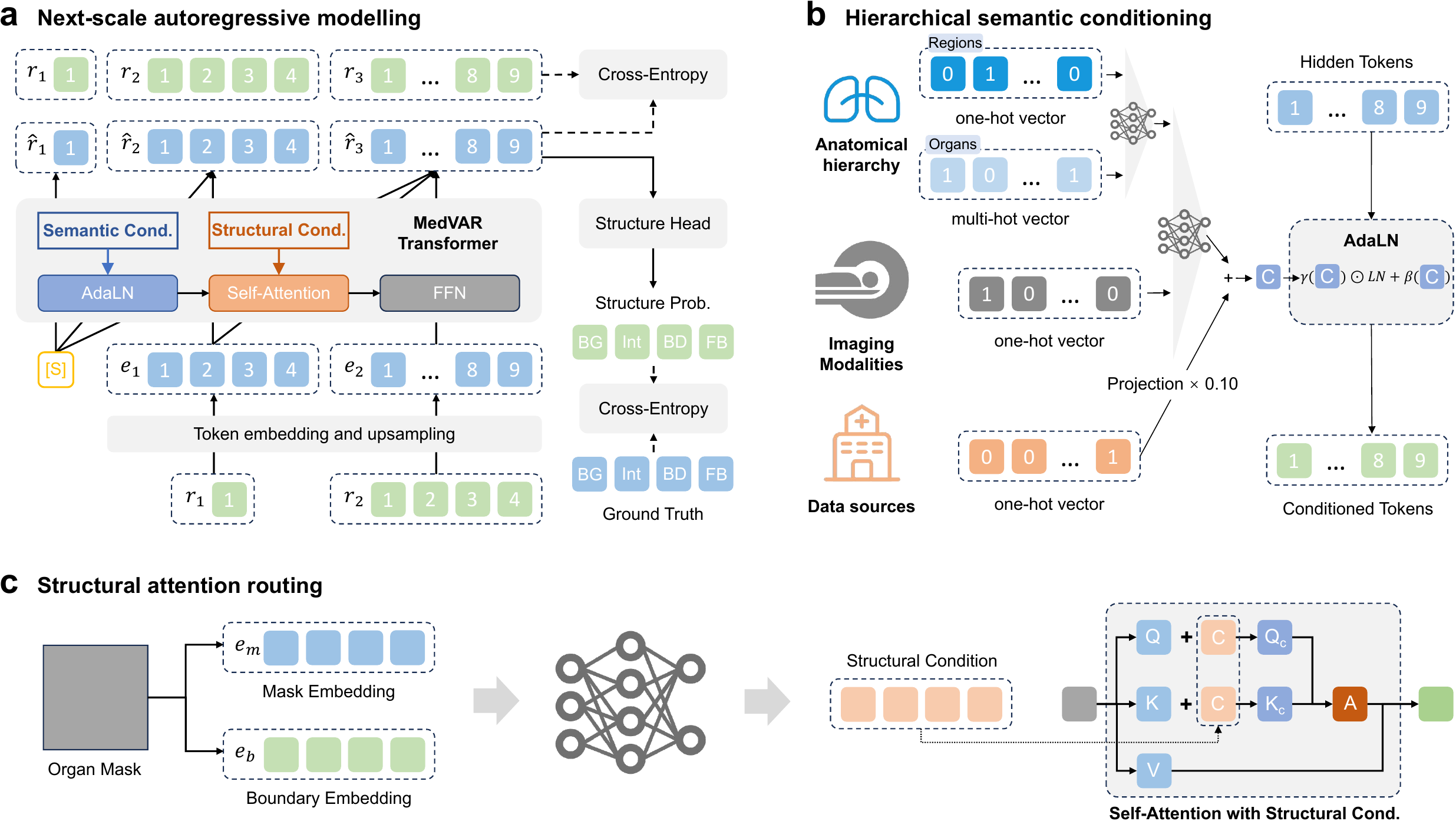}
    \caption{\figuresixcaption}
    \label{fig:method}
\end{figure*}

\subsection{Generative baselines and fidelity evaluation}

The comparison covered adversarial, denoising diffusion, diffusion transformer and discrete autoregressive generators. We trained StyleGAN3 \cite{karras2021alias}, DDPM \cite{ho2020ddpm}, DiT \cite{peebles2023scalable}, MaskGIT \cite{chang2022maskgit} and RQ-Transformer \cite{lee2022residual} as baselines. They used the same preprocessed images and data source labels as MedVAR, with labels supplied through each architecture's native conditioning interface. DDPM-S and DiT-S each contained approximately 330 million parameters, whereas DDPM-L and DiT-L each contained approximately 600 million parameters. These configurations were matched approximately to depth-16 and depth-20 MedVAR by parameter count. The DDPM and DiT were separately evaluated at 10, 20, 30, 50 and 100 sampling steps to characterize the quality--latency trade-off at a fixed model size. MedVAR was evaluated at all four model sizes with semantic conditions alone and with both semantic and structural conditions. The comparison across generator families controlled the training images, requested conditions, reference images, sample count and evaluator, while retaining the sampling procedure specified for each architecture. Additional medical generators were HA-GAN \cite{sun2022hierarchical}, WDM \cite{friedrich2024wdm}, MOTFM \cite{yazdani2025flow} and the evaluated CT and MRI configurations of MAISI \cite{guo2025maisi}. FLUX.1-dev \cite{flux1dev2025}, Stable Diffusion 3.5 Medium \cite{sd35medium2024} and PRX-1024-T2I-Beta \cite{prx1024t2i2025} provided zero-shot general text-to-image references. Qualitative comparisons included MaskGIT, RQ-Transformer, StyleGAN3, and the largest evaluated DDPM and DiT configurations (DDPM-L and DiT-L), alongside MedVAR and real reference images.

FID was computed from Inception features \cite{heusel2017gans}. Fr\'echet distance computed with RadImageNet features is reported as RadFID \cite{woodland2024_fid_med}. KID used the polynomial kernel estimator \cite{binkowski2018demystifying}. CLIP maximum mean discrepancy (CMMD) used the published estimator in CLIP feature space \cite{jayasumana2024rethinkingfidbetterevaluation}. Within each quantitative analysis, generators used the same evaluator, fixed reference images and sample count. Generated and reference images were compared for the same requested conditions and were not required to be paired. For the comparison across generator families in Fig.~2, FID used ten bootstrap replicates, each based on a 25,000-image resample, and their mean was reported. The matrix in Fig.~2c ranked the available FID values separately within each anatomy--modality group; unavailable combinations were marked N.A. and excluded from that group's ranking. Its MedVAR row used the depth-30 model with semantic conditions alone under the main sampling protocol. FID was also computed separately for every non-empty anatomy--modality group. Let $\mathcal G$ contain the non-empty anatomy--modality groups, with FID $F_g$ and $n_g$ real reference images for group $g$. The subset $\mathcal G_{\leq1000}$ contains groups with at most 1,000 real images. The three group summaries were
\begin{align}
F_{\mathrm{weighted}}&=\frac{\sum_{g\in\mathcal G}n_gF_g}{\sum_{g\in\mathcal G}n_g},\\
F_{\mathrm{macro}}&=\frac{1}{|\mathcal G|}\sum_{g\in\mathcal G}F_g,\\
F_{\mathrm{tail}}&=\frac{1}{|\mathcal G_{\leq1000}|}
\sum_{g\in\mathcal G_{\leq1000}}F_g.
\end{align}

The overview in Fig.~1e used depth-30 MedVAR with semantic conditions alone as the reference. It summarized nine FID values for anatomy--modality groups and one descriptive measure of quality and efficiency. For method $j$, the latter was $E_j=F_j[\log_{10}(1+P_j)]^{0.1}$, where $F_j$ is FID and $P_j$ is sampling latency in seconds for a batch size of one. Each lower-is-better measure $H$ was expressed relative to the corresponding MedVAR value and mapped to the displayed score as
\begin{equation}
r_{j,H}=\frac{H_{\mathrm{MedVAR}}}{H_j},\qquad
S_{j,H}=\frac{2r_{j,H}}{1+r_{j,H}}.
\end{equation}
MedVAR therefore equals 1; values below this reference denote a higher raw value than MedVAR, and values above it denote a lower raw value. The quality--efficiency quantity was included only as a compact visual summary; the primary efficiency comparison is the relation between FID and latency in Fig.~2d.

\looseness=-1
Generation diversity was assessed within matched conditions using multiscale structural similarity (MS-SSIM) \cite{wang2003multiscale} and Learned Perceptual Image Patch Similarity (LPIPS) \cite{zhang2018perceptual}. Lower MS-SSIM and higher LPIPS indicate greater variation. Values were interpreted relative to held-out real images and jointly with fidelity, because diversity alone can increase through anatomically invalid variation.

\subsection{Condition recognizability and classification accuracy score}

Condition recognizability was measured with a nine-class anatomy--modality classifier trained on real images and evaluated on generated images. Classification accuracy score (CAS) \cite{ravuri2019classification} used the reciprocal training direction: a classifier was trained on generated images and evaluated on held-out real images. The classes were abdomen CT, abdomen MRI, brain MRI, chest CT, heart CT, heart MRI, prostate MRI, spine CT and spine MRI. Balanced accuracy was the mean recall across the nine classes, with a chance value of $1/9$. Both directions used a RadImageNet-pretrained ResNet-50 at $224\times224$ resolution. Models were trained for 30 epochs with class-balanced sampling, AdamW (learning rate $3\times10^{-4}$ and weight decay $10^{-4}$), batch size 96 and seed 6201. The checkpoint maximizing validation balanced accuracy was used for evaluation. The classifier trained on real images was also evaluated on the held-out real images to provide an upper reference for recognizability. Figure~2f reports balanced accuracy with 95\% confidence intervals from 1,000 class-stratified bootstrap samples. All generators used identical class definitions, image preprocessing and held-out test data.

\subsection{Multicentre HCC lesion segmentation}

The HCC experiment evaluated whether images sampled from a fixed pretrained generator could support representation learning before lesion segmentation with limited labelled data. The HCC image distribution contributed to generator pretraining, after which the generator was frozen. Downstream training, validation and test partitions were patient-disjoint and fixed across all arms. Within the training partition, 40\% of cases from every centre were selected independently for seeds 6201--6203, giving 307 labelled volumes and 1,526--1,570 lesion-positive slices per seed. The validation and test partitions contained 108 and 216 volumes, respectively. No patient identifier, source image or lesion mask was supplied during generation. The generator produced 420 images using abdomen, liver, MRI and centre labels (60 images per centre), and 420 unconditional images with all learned condition embeddings set to their null indices.

All three pretraining groups used the same four-level two-dimensional U-Net \cite{ronneberger2015unet} with base width 32 at $256\times256$ resolution. Self-supervision used masked image reconstruction. Random $16\times16$ blocks covering 75\% of each image were replaced with uniform noise. The loss was L1 error on masked pixels plus 0.05 times L1 error on visible pixels. In the first stage, the control used 420 real images per epoch, whereas each of the other two groups used 420 generated images per epoch. This stage ran for 30 epochs with batch size 16 and AdamW (learning rate $10^{-3}$ and weight decay $10^{-4}$). All groups then underwent the same 15-epoch second stage on real images, using 1,680 images sampled with replacement per epoch and a learning rate of $5\times10^{-4}$. The segmentation output layer was reset after this stage so that the downstream comparison transferred the learned representation rather than the output layer used for image reconstruction.

The pretrained U-Net was fine-tuned for 80 epochs using batch size 12 and AdamW (learning rate $10^{-3}$ and weight decay $10^{-4}$). The loss was soft Dice plus 0.5 times weighted binary cross-entropy. The positive-class weight was estimated from real training masks and capped at 25. Training used sampling balanced across centres and cases and standard online geometric and intensity augmentation. Predictions were thresholded at 0.5, and the checkpoint maximizing validation mean volume Dice was evaluated on held-out real volumes. Dice was computed per volume, averaged within each centre and then macro-averaged across the seven centres. All comparisons used three matched seeds.

The feature analysis used one central slice from each of 60 real cases per centre. Comparators were 60 images generated with each centre label and 420 unconditional images. A RadImageNet-pretrained ResNet-50 produced features after global average pooling at $224\times224$ resolution. Features were standardized using the real subset. Principal component analysis (PCA) was fitted to real features, and t-distributed stochastic neighbour embedding (t-SNE) was fitted to the joint feature set. Both were displayed in two and three dimensions. Contours and centroids were included as visual guides. A class-balanced logistic-regression classifier ($C=2$) was trained on 70\% of the sampled real cases and tested on the remaining 30\%. Accuracy for the centre labels was the fraction of generated images assigned to the label supplied during generation. Maximum mean discrepancy (MMD) \cite{gretton2012kernel} between real and generated images used a radial-basis-function kernel with the median positive pairwise squared distance as bandwidth. Accuracy and MMD for each centre were reported as descriptive measures.

\subsection{Through-plane reconstruction and frozen 3D evaluation}

For through-plane reconstruction, an interior target slice $x_0$ was withheld and its immediately preceding and following slices, $x_-$ and $x_+$, were provided as the only image observations. Their arithmetic mean, $x_{\mathrm{lin}}=(x_-+x_+)/2$, defined the linear-interpolation baseline. The CT split contained 80 training, 24 validation and 40 held-out volumes; the MRI split contained 20 training, four validation and 40 held-out volumes. The held-out analyses comprised 3,223 CT and 525 MRI target slices. The MRI test cohort included abdominal MRI and prostate MRI from six acquisition centres.

The frozen MedVAR model $G_{\theta}$ received $x_{\mathrm{lin}}$ as its image input and generated a deterministic image $x_{\mathrm M}=G_{\theta}(x_{\mathrm{lin}};c)$ under the corresponding semantic condition $c$. The withheld target served as the reconstruction label and evaluation reference. MedVAR generation used classifier-free guidance 2.0, top-$k=1$ and top-$p=1.0$. A multiscale U-Net $R_{\psi}$ received the two observed slices, the MedVAR image and the difference between the MedVAR image and linear interpolation. It fused these inputs at four encoder resolutions and reconstructed the target as
\begin{equation}
\widehat{x}_{\mathrm{MedVAR}}
=x_{\mathrm{lin}}+0.25\tanh R_{\psi}(x_-,x_+,x_{\mathrm M},x_{\mathrm M}-x_{\mathrm{lin}}).
\end{equation}
The base widths were 48 for CT and 32 for MRI. Three networks were trained for each modality with seeds 6201--6203. Checkpoints were selected by validation mean squared error. An independent residual U-Net $U_{\omega}$ \cite{ronneberger2015unet} with base width 32 received only the observed slices and used the matched output formulation,
\begin{equation}
\widehat{x}_{\mathrm{U\mbox{-}Net}}
=x_{\mathrm{lin}}+0.25\tanh U_{\omega}(x_-,x_+).
\end{equation}
This baseline was evaluated independently and was not part of the system using MedVAR. Reconstruction networks were optimized with AdamW at $256\times256$ resolution using weighted L1, mean squared error and image gradient losses. CT networks were trained for up to 100 epochs with a cosine learning rate schedule from $8\times10^{-4}$ to $10^{-5}$. MRI networks were trained for 60 epochs at a learning rate of $10^{-3}$. PSNR was computed on complete reconstructed slices and averaged first within each held-out volume. Mean squared error selected validation checkpoints. With intensities fixed to $[0,1]$, PSNR is derived from this error and was reported as the test outcome. Uncertainty was estimated with 20,000 bootstrap resamples across volumes and training seeds. These comparisons evaluate the complete reconstruction systems rather than the contributions of individual inputs or network capacity.

For volumetric evaluation, a three-dimensional full-resolution nnU-Net was trained for 100 epochs on 40 real densely sampled CT volumes and their binary foreground-union masks. The evaluator was frozen before reconstruction scoring. We removed a deterministic subset of interior slices from 36 held-out volumes; the missing fraction ranged from 33.3\% to 49.2\%. The missing slices were reconstructed by linear interpolation, the residual U-Net or the reconstruction network using MedVAR images. The same fixed evaluator scored every reconstructed stack and the corresponding fully sampled volume. Three-dimensional Dice was computed against the same real foreground-union mask for every input variant, with comparisons paired within volume. Confidence intervals used 20,000 bootstrap resamples across volumes and training seeds. Fully sampled inputs were used as the reference.

\subsection{White-box membership inference}

The membership analysis used the depth-16 checkpoint (Fig.~5a, b). The white-box adversary evaluated MedVAR by teacher forcing and read the probability assigned to every ground-truth image token. The attack tested whether an individual slice had contributed to training, rather than whether any image from the same patient had contributed. The adversary knew the semantic conditions supplied during training but not the structural condition; the exact mask--boundary input was evaluated separately as a setting with access to additional information. To reduce anatomical confounding, each member slice was paired with a nonmember slice from the same case at a nearby through-plane position, forming 43,828 pairs (87,656 images). Cases, rather than slices, were assigned to mutually exclusive calibration and test partitions using a fixed hash of the case identifier. The calibration partition contained 2,388 cases and 44,174 image rows, and the test partition contained 2,364 cases and 43,482 rows. This analysis of individual-slice membership is distinct from the independent patient-disjoint candidate cohort used for nearest-neighbour copy detection below.

For a query image $x$, the frozen tokenizer produced its ground-truth coarse-to-fine sequence $v_{1:T}$, where $T=680$. Let $z^{c}_{t,j}$ and $z^{0}_{t,j}$ denote the teacher-forced logits for codebook entry $j$ at position $t$ under the known semantic conditions and the null condition used during training, respectively. Conditional token loss and the contrastive CFG token loss were
\begin{equation}
L^{c}_{t}=-\log \operatorname{softmax}(z^{c}_{t})_{v_t},\qquad
L^{\mathrm{CFG}}_{t}=-\log \operatorname{softmax}(z^{c}_{t}-z^{0}_{t})_{v_t}.
\end{equation}
These contrastive logits follow the image-autoregressive attack protocol and are unrelated to the CFG scale used for sampling \cite{kowalczuk2025privacy}.

We recomputed seven published likelihood-based score families on this common target model: conditional loss \cite{yeom2018privacy}, CFG loss and hinge score \cite{kowalczuk2025privacy}, zlib-normalized loss \cite{carlini2021extracting}, Min-K \cite{shi2024detecting}, Min-K++ \cite{zhang2025minkpp} and context-aware membership inference attack (CAMIA) \cite{chang2025context}. Writing $\ell_t=-L^{\mathrm{CFG}}_t$, $p_{t,j}=\operatorname{softmax}(z^{c}_{t}-z^{0}_{t})_j$ and $m_t=(z^{c}_{t,v_t}-z^{0}_{t,v_t})-\max_{j\ne v_t}(z^{c}_{t,j}-z^{0}_{t,j})$, the main oriented scores included
\begin{equation}
\begin{aligned}
s_{\mathrm{cond}}&=-T^{-1}\sum_t L^{c}_t,&
s_{\mathrm{CFG}}&=-T^{-1}\sum_t L^{\mathrm{CFG}}_t,\\
s_{\mathrm{zlib}}&=-\frac{T^{-1}\sum_t L^{\mathrm{CFG}}_t}{Z(v_{1:T})},&
s_{\mathrm{hinge}}&=T^{-1}\sum_t m_t,\\
s_{\mathrm{MinK}}&=-|I_K|^{-1}\sum_{t\in I_K}L^{\mathrm{CFG}}_t,&
s_{\mathrm{MinK++}}&=|I^{+}_K|^{-1}\sum_{t\in I^{+}_K}q_t.
\end{aligned}
\end{equation}
Here $Z(\cdot)$ is the zlib-compressed byte length, $I_K$ indexes the $K\%$ largest token losses, and $I^{+}_K$ indexes the $K\%$ smallest standardized target log-probabilities. The latter were $q_t=(\ell_t-\mu_t)/(V_t+10^{-6})$, with $\mu_t=\sum_jp_{t,j}\log p_{t,j}$ and $V_t=\sum_jp_{t,j}(\log p_{t,j})^2-\mu_t^2$, matching the evaluated implementation. Candidate values $K\in\{10,20,30,40,50\}$ were ranked on calibration TPR at 1\% FPR and then calibration ROC AUC, selecting Min-K 40\% and Min-K++ 50\%. Five CAMIA sequence statistics were evaluated by the same calibration-only rule; the selected statistic was the negative fraction of $q_t$ values below their within-image mean. No test labels were used for method selection.

Four additional scores adapted this token evidence to the heterogeneous medical setting. First, conditional loss was standardized within strata defined by dataset and mask availability, $g=(\text{dataset},\text{mask availability})$, using nonmembers in the calibration set:
\begin{equation}
s_{\mathrm{strat}}(x)=\frac{\mu_g-\overline{L}^{c}(x)}{\max(\sigma_g,10^{-8})}.
\end{equation}
Estimates from the corresponding stratum were used when at least 50 calibration nonmembers were available and were otherwise replaced by estimates from all calibration nonmembers. Second, the multiscale logistic-regression score used
\begin{equation}
s_{\mathrm{multi}}(x)=\sigma\!\left(\beta_0+\boldsymbol{\beta}^{\mathsf T}\widetilde{\mathbf f}(x)\right),
\end{equation}
where $\mathbf f$ contained mean CFG loss at each of the ten token-map scales together with global CFG loss, hinge loss, Min-K 20\% and Min-K++ 20\%. Features were standardized and the coefficients were fitted only on the calibration partition using class-balanced, L2-regularized logistic regression ($C=1$). Third, with $Q_{0.95}$ denoting the 95th percentile of all conditional token losses from calibration nonmembers, the score based on high-loss tokens was $s_{\mathrm{tail}}(x)=-T^{-1}\sum_t\mathbf{1}\{L^c_t>Q_{0.95}\}$. Fourth, the loss obtained after providing the exact mask and boundary was $s_{\mathrm{struct}}(x)=-T^{-1}\sum_tL^{c,d}_t$, where $L^{c,d}_t$ was recomputed with this additional input. This score represents a setting with access to additional information and was not part of the primary threat model.

\looseness=-1
Controls using only metadata or image intensities contained no MedVAR likelihood. The metadata control combined dataset, data source, modality, region, organ labels, mask availability, slice position and foreground fraction; the image control used standardized $16\times16$ pooled grayscale intensities. Both were class-balanced logistic regressions fitted on calibration cases only and quantified membership separability caused by cohort or image-distribution differences \cite{bai2026splitbias}. All scores were oriented so that larger values indicated greater evidence of membership. Attack classifier coefficients, score variants and operating thresholds were fitted or selected on the calibration partition and then held fixed for test evaluation.

For a threshold $\tau$, with membership label $y_i\in\{0,1\}$ and oriented attack score $s_i$, we calculated
\begin{equation}
\begin{aligned}
\operatorname{TPR}(\tau)&=\frac{\sum_i\mathbf{1}\{y_i=1,\ s_i\geq\tau\}}{\sum_i\mathbf{1}\{y_i=1\}},\\
\operatorname{FPR}(\tau)&=\frac{\sum_i\mathbf{1}\{y_i=0,\ s_i\geq\tau\}}{\sum_i\mathbf{1}\{y_i=0\}}.
\end{aligned}
\end{equation}
Here $\mathbf{1}\{\cdot\}$ denotes the binary indicator. ROC AUC summarizes ranking over all thresholds; its chance value is 0.5. TPR at 1\% FPR measures recovery under a stringent false-alarm constraint and has a random reference of 1\%. The nominal 1\% FPR threshold was the 99th percentile of scores from calibration nonmembers, $\tau_{0.01}=Q_{0.99}\{s_i:y_i=0,\ i\in\mathcal{C}\}$. Ninety-five per cent bootstrap percentile confidence intervals were obtained from 1,000 samples clustered by case, thereby retaining correlations among slices from the same case.

Figure~5b presents the low-FPR region using the same test labels and attack scores as Fig.~5a. Both ROC axes span $10^{-4}$ to $10^{0}$ on logarithmic scales; zero-valued ROC coordinates, which are undefined under a logarithm, were omitted. The diagonal $\operatorname{TPR}=\operatorname{FPR}$ denotes random guessing, and the vertical line marks the $\operatorname{FPR}=10^{-2}$ operating point \cite{carlini2022membership}.

\subsection{Nearest-neighbour copy detection}

Copy detection was evaluated independently of token likelihoods using a nearest-neighbour embedding analysis based on medical memorization studies \cite{dar2026memorization}. The cohort contained 4,705 paired cases and 86,922 image candidates, with training and held-out test candidates kept patient-disjoint. The evaluated generated sets came from the depth-16 checkpoint under the fixed sampling protocol used for the privacy analysis. Three ResNet-50 encoders were initialized from ImageNet weights and trained independently for 50 epochs on real training members only. Training used a 128-dimensional contrastive objective and batches containing images from one dataset at a time. The encoders were frozen before scoring. Exact copies and small rotations, flips and contrast changes were used as positive calibration controls. For candidate real image $x$, generated set $\mathcal{G}$ and encoder $f$, the attack score was the largest cosine similarity to any generated image,
\begin{equation}
s_{\mathrm{NN}}(x;\mathcal{G})=\max_{g\in\mathcal{G}}
\frac{f(x)^{\mathsf{T}}f(g)}{\lVert f(x)\rVert_2\lVert f(g)\rVert_2}.
\end{equation}
A candidate was therefore considered more likely to be a member when at least one generated image lay unusually close to it in feature space. The numbers of generated images were fixed at 0.5, 1 and 2 times the size of the training set. ROC AUC and TPR at 1\% FPR were calculated for every combination of encoder and set size. Uncertainty within each combination used 1,000 bootstrap samples. Figure~5c reports individual encoders with their mean and sample standard deviation. Chance references were 0.5 for ROC AUC and 1\% for TPR at 1\% FPR.

For qualitative comparison, cases in the $2\times$ generated pool were stratified by mean similarity to training images across the three encoders and sampled across datasets and modalities. Each generated image was paired separately with its nearest training image and nearest held-out test image. Normalized cross-correlation and structural similarity \cite{wang2004ssim} were reported next to their corresponding image pair. Within each row, the three unique grayscale source images shared a display window defined by their pooled 0.5th and 99.5th percentiles. Absolute differences from the training and test neighbours shared a magma colour scale within each row. The scale was capped at the pooled 99.5th percentile of both difference distributions. The two pairs are therefore comparable within a row, whereas colours in the difference maps are not absolute measurements across rows.

\subsection{Testing data source labels under simulated contamination}

A $2\times2$ factorial stress test varied whether a label identifying the data source was provided and whether simulated contamination was added to the training set. The contaminant comprised deterministic, independent standard Gaussian images clipped to $[-1,1]$ and was used only during training. Its sample count equalled the median number of images among the real training sources. Each contaminant image inherited the anatomical region, organ and modality labels of a randomly sampled real training image but received a distinct source label; validation and reference images remained uncontaminated. This design isolates the contribution of the source label while retaining the distribution of the other semantic conditions. It follows the motivation for providing source labels in autoregressive imaging across multiple sites \cite{nezhad2025fedgat}. Every group used the depth-16 MedVAR model with CFG 4.5, top-$k$ 1,300, top-$p$ 0.98 and generation seed 42. FID \cite{heusel2017gans} was computed relative to reference images from AMOS ($n=4,112$), AbdomenCT-1K ($n=11,161$), BTCV ($n=218$) and WORD ($n=2,366$). For each source, the interaction was defined as
\begin{equation}
\begin{aligned}
\Delta \mathrm{FID}_{\mathrm{interaction}}={}&
(\mathrm{FID}_{\mathrm{off,contaminated}}-\mathrm{FID}_{\mathrm{off,clean}})\\
&-(\mathrm{FID}_{\mathrm{on,contaminated}}-\mathrm{FID}_{\mathrm{on,clean}}).
\end{aligned}
\end{equation}
A positive interaction indicates that providing the source label attenuated the FID change associated with contamination; a negative value indicates a response in the opposite direction.

\subsection{Statistical analysis}

HCC segmentation results are reported as the mean and sample standard deviation across three matched training seeds. In Fig.~4b, bars show each centre's mean paired Dice gain and points show individual seeds. The analyses of image features, classification of centre labels and MMD report descriptive point estimates. Reconstruction comparisons were paired at the image or volume level. Confidence intervals for three-dimensional Dice used 20,000 bootstrap resamples across volumes and training seeds. Privacy confidence intervals used 1,000 bootstrap samples clustered by case. Nearest-neighbour estimates from each encoder were summarized by their mean and sample standard deviation. The experiment with source labels and simulated contamination reports FID for a fixed generation seed; analyses of image quality and the ablations report point estimates. Representative images comparing generator families, model sizes, diversity and controllability were matched by anatomy, view and condition. Reconstruction and nearest-neighbour examples followed the selection criteria described in their respective Methods sections.